\documentclass[11pt,a4paper]{article}
\usepackage{authblk}
\usepackage{blindtext}
\usepackage[hyperref]{acl2018}
\usepackage{times}
\usepackage{latexsym}
\usepackage{graphicx}
\usepackage{multirow}
\usepackage{multicol}
\usepackage{enumitem}
\usepackage{url}
\usepackage{mdwlist}
\usepackage{theorem}
\theoremstyle{break}
\newtheorem{example}{Example}

\aclfinalcopy 

\usepackage{etoolbox}
\makeatletter
\patchcmd{\maketitle}
{\def\@makefnmark}
{\def\@makefnmark{}\def\useless@macro}
{}{}
\makeatother

\title{Adaptations of ROUGE and BLEU to Better Evaluate Machine Reading Comprehension Task}

\author[1 2*]{An Yang\thanks{\llap{\textsuperscript{*}}This work was done while the first author was doing internship at Baidu Inc.}}
\author[2]{Kai Liu}
\author[2]{Jing Liu}
\author[2]{Yajuan Lyu}
\author[1]{Sujian Li}
\affil[1]{Key Laboratory of Computational Linguistics, Peking University, MOE, China}
\affil[2]{Baidu Inc., Beijing, China}
\affil[ ]{\tt {\{yangan, lisujian\}@pku.edu.cn}}
\affil[ ]{\tt {\{liukai20, liujing46, lvyajuan\}@baidu.com}}

\date{}

\begin{document}
\maketitle
\begin{abstract}
  Current evaluation metrics to question answering based machine reading comprehension (MRC) systems generally focus on the lexical overlap between candidate and reference answers, such as ROUGE and BLEU.
  However, bias may appear when these metrics are used for specific question types, especially questions inquiring yes-no opinions and entity lists.
  In this paper, we make adaptations on the metrics to better correlate $n$-gram overlap with the human judgment for answers to these two question types.
  Statistical analysis proves the effectiveness of our approach.
  Our adaptations may provide positive guidance for the development of real-scene MRC systems.
\end{abstract}

\section{Introduction}

The goal of current MRC tasks is to develop agents which are able to comprehend passages automatically and answer open-domain questions correctly.
With the release of several large-scale datasets like SQuAD \cite{rajpurkar2016squad}, MS-MARCO \cite{nguyen2016ms} and DuReader \cite{he2017dureader}, many MRC models have been proposed in previous works \cite{wang2016machine, seo2016bidirectional, wang2017gated}. 
Although MRC model architectures have been intensively studied, the evaluation metrics for them are rarely discussed. 
For early cloze-style and multiple choice datasets \cite{richardson2013mctest, hermann2015teaching}, this may not be problematic. 
However, considering the trend that the model is required to generate answers and question type is becoming more variable and closer to real cases, we believe the design of evaluation metric is indeed an issue to be focused on.

Currently, the criterion for comparing generated and gold answers is mostly based on lexical overlap.
For example, SQuAD uses exact-match ratio and word-level F1-score, while MS-MARCO and DuReader employ ROUGE-L \cite{lin2004rouge} and BLEU \cite{papineni2002bleu} which measure $n$-gram consistency or longest common sequence (LCS) length.
For some question types, we notice these metrics may not correlate with semantic correspondence well in some cases.
In this paper, we mainly tackle the issue of yes-no and entity questions.
For yes-no questions, overlap-based metrics may ignore the yes-or-no opinion which is more crucial in determining agreement between answers. 
Answers with contrary opinions may have high lexical overlap, such as ``\textit{The radiation of wireless routers has an impact on people}'' and ``\textit{The radiation of wireless routers has no impact on people}''. 
Similarly, for entity questions, we think the agreement should be more reflected by the correctness of entity listing. 
Answers which lack or mispredict entities should be in distinction from correct answers, but the mistakes actually affect little in BLEU and ROUGE, especially when the entity is a number.
These two question types are quite common in MRC datasets and real scenario.
As is shown in \newcite{he2017dureader}, 36.2\% queries in DuReader and 47.5\% in Baidu real search data are classified into these two categories.
For the reasons above, developing an automatic evaluation system which takes consideration of the inherent characteristics of these question types is of great necessity.

In previous work, \newcite{dang2007overview} employed type-specific metrics for evaluating candidate answers in TREC 2007 QA track. 
Setting the accuracy of yes-no opinion type and F1-score of entity list as extra metrics may solve the problem to some extent. 
However, from the perspective of simplicity and scalability to growing question type category, we hope to design a unified and end-to-end evaluation metric which is calculated automatically.
We propose some adaptations for ROUGE and BLEU which provide them awareness to yes-no opinion and entity agreement.
Compared with original metrics, our modified ROUGE and BLEU achieve higher correlation to human judgment on DuReader samples in both type-specific and overall analysis.
Our work is a preliminary exploration of better automatic evaluation systems for MRC model in the real application.

In the remainder of this paper, related work is discussed in section 2. Then we give details about our adaptation on ROUGE and BLEU in section 3. Statistical analysis is given in section 4. In section 5, we conclude the paper.

\section{Related Work}

\paragraph{MRC Task} 
Recent years have witnessed growing research interest in machine reading comprehension. 
Annotation of large-scale datasets is a strong driving force for the recent progress of MRC systems.
The paradigm of such datasets ranges from cloze test \cite{hermann2015teaching, hill2015goldilocks}, multiple choice \cite{lai2017race}, span extraction  \cite{rajpurkar2016squad} and answer generation \cite{nguyen2016ms, he2017dureader}. 
The last paradigm with multi-passages and manually annotated answers for each question is more close to real application.
Based on these resources, end-to-end neural MRC model architectures are implemented, including match-LSTM \cite{wang2016machine}, BiDAF \cite{seo2016bidirectional}, DCN \cite{xiong2016dynamic} and r-net \cite{wang2017gated}.
With the objective of lexical overlap based evaluation metrics, these models focus more on text matching to references, which has bias to human demand.
Instead, conceiving opinion and entity-aware metrics will encourage future MRC systems to look more into real application cases.

\paragraph{QA Evaluation Metrics} 
In the past competitions on question answering, various evaluation metrics were proposed to make comparisons between participating systems.
Early tasks including TREC-8 and TREC-9 QA tracks \cite{voorhees1999trec, voorhees2000overview} only consist of factoid questions. 
The ordered candidate answers are evaluated manually to give binary correctness judgment and summarized by mean reciprocal rank (MRR). 
With the addition of complex non-factoid questions such as definition questions in TREC 2003 \cite{voorhees2003overview} and ``other'' questions in TREC 2007 \cite{dang2007overview}, manual assessment becomes more difficult.
``Nugget pyramids'' \cite{nenkova2007pyramid} are employed for scoring, which prefer answers coveraging more key points (nuggets).
The nuggets are annotated and weighted by human assessors, which is labor-intensive.
\newcite{breck2000evaluate} proposed to use word recall against the stemmed gold answer as an automatic evaluation metric.
Following this idea, metrics evaluating $n$-gram overlap and LCS length between candidate and gold answers are designed and become prevalent, among which BLEU \cite{papineni2002bleu} and ROUGE \cite{dang2007overview} are most widely-used.
In general, BLEU focuses more on $n$-gram precision and ROUGE is recall-oriented. 
Later work has made adaptations on these metrics from different perspectives  \cite{banerjee2005meteor, liu2008correlation}.
In this paper, our adaptations are aimed at increasing their correlation to real human judgment on yes-no and entity question answering, which are proved to be practical.

\section{Methodology}

The brief idea of our adaptations is to add additional lexical overlap items which can reflect opinion and entity agreement as the bonus. 
In the official evaluation of MS-MARCO and DuReader, ROUGE-L and BLEU are employed as metrics at the same time, with the former as the primary criterion for ranking participating systems. 
Their modifications will be elaborated separately.

\subsection{Adaptations on BLEU}

For one question sample with single candidate and several gold answers, \newcite{papineni2002bleu} define cumulative BLEU-n with uniform $n$-gram weight as follows:
\begin{equation} 
	BLEU_{cum} = BP \cdot \left( \prod_{i=1}^n{P_i} \right) ^ {\frac{1}{n}}
\end{equation}

In the equation, $P_i$ is the precision of $i$-gram in the candidate answer
\begin{equation} 
	P_i = \frac{\sum\limits_{i \textendash gram \in \mathcal{C}}{Count_{clip} \left( i \textendash gram \right)}}{\sum\limits_{i \textendash gram' \in \mathcal{C}}{Count \left( i \textendash gram' \right)}}
\end{equation}
where $\mathcal{C}$ is $i$-gram set of the candidate answer, $Count(x)$ calculates the number of times that $i$-gram $x$ appear in candidate and $Count_{clip}(x)$ clips $Count(x)$ to the maximum times that $x$ appears in references.

$BP$ stands for brevity penalty item, given reference length $r$ and candidate length $c$
\begin{equation} 
BP = e^{\min{(1-\frac{r}{c}, 0)}}
\end{equation}
For cases with mutiple reference answers, we choose the reference length which is closest to $c$.

For yes-no questions, we add an additional term into both the numerator and denominator of (2) to measure yes-no opinion agreement
\begin{equation} 
	bonus_{yn} = \alpha\sum\limits_{i \textendash gram \in \mathcal{C}}{Count_{clip \textendash s} \left( i \textendash gram \right)}
\end{equation}
where $Count_{clip \textendash s}(x)$ clips $Count(x)$ to the maximum times that $x$ appears in reference answers sharing the \textbf{same} yes or no opinion with the candidate and $\alpha$ stands for bonus weight. 
If the participant correctly judges the opinion type, its adapted BLEU score will increase due to the introduced bonus. However, the BLEU score still never exceed 1.0.

Calculating $bonus_{yn}$ requires the opinion labels of both candidate and gold answers. 
For references, it does not consume much labor to annotate opinion labels in the construction of datasets. 
We notice that recent DuReader dataset already satisfies this requirement, with each yes-no reference answer labeled ``Yes'', ``No'' or ``Depends''. 
For candidate answers, we think it should be the trend to encourage participating systems to provide explicit predicted opinion labels apart from the answers. The following example gives a simple illustration of how to compute $P_i$ with consideration of $bonus_{yn}$.

{\small \begin{example}[Adapted $P_2$ for yes-no answer]
	\label{exam1}
	\textbf{Question}: Is skipping rope an aerobic exercise?
		
	\noindent \textbf{Predicted answer}: [Yes] Skipping rope is an aerobic exercise. 
	
	\noindent \textbf{Gold answer 1}: [Yes] Skipping rope is a kind of aerobic exercise with low intensity.
	
	\noindent \textbf{Gold answer 2}: [Depends] Skipping rope can be regarded as an aerobic exercise only when skipping for a long time.
	
	\noindent \textbf{Number of predicted bigrams\footnote{Include period symbol and omit lemmatization.}}: 6
	
	\noindent \textbf{Number of hit predicted bigrams}: 4
	
	\noindent \textbf{Bigram count for bonus}: 3 (hit gold answer 1)
	
	\noindent \textbf{Vanilla $P_2$}: 4 / 6 = 0.67
	
	\noindent \textbf{Adapted $P_2$ ($\alpha=1.0$)}: (4 + 3) / (6 + 3) = 0.78

\end{example}}

Similarly, we add another term to the numerator and denominator of (2) for bonusing correct entity answers
\begin{equation} 
	bonus_{ent} = \beta\sum\limits_{i \textendash gram \in \mathcal{C}}{Count_{clip \textendash e} \left( i \textendash gram \right)}
\end{equation}
where the reference answers provide a gold entity list and $Count_{clip \textendash e}(x)$ clips $Count(x)$ to the maximum times that $x$ appears in the entity strings in the list. 
$\beta$ stands for the weight of entity bonus. 
As a result, the score of answer containing more right entities will increase, as is shown in example~\ref{exam2}.

{\small \begin{example}[Adapted $P_2$ for entity answer]
	\label{exam2}
	\textbf{Question}: How long did it take for Qin Dynasty to unify China?

	\noindent \textbf{Predicted answer}: Qin unified China in 221 BC after the war against other kingdoms which lasted ten years.
	
	\noindent \textbf{Gold answer}: Qin unified China in ten years, from 230 BC to 221 BC.
	
	\noindent \textbf{Gold entity set}: ten years, 230 BC, 221 BC
	
	\noindent \textbf{Number of predicted bigrams}: 16
	
	\noindent \textbf{Number of hit predicted bigrams}: 5
	
	\noindent \textbf{Vanilla $P_2$}: 5 / 16 = 0.31
	
	\noindent \textbf{Adapted $P_2$ ($\beta=1.0$)}: (5 + 2) / (16 + 2) = 0.39
	
\end{example}}

To calculate BLEU score over entire dataset using (1), we follow the common approach to compute overall $P_i$, which separately sums the numerator and denominator of (2) with bonus terms over all the samples and finally get them divided. The $r$ and $c$ for $BP$ are also the sum across whole dataset.

\subsection{Adaptations on ROUGE-L}

As mentioned in \newcite{lin2004rouge}, the principle of calculating ROUGE-L is to examine the precision and recall between candidate and reference answers considering longest common subsequences. For single sample, ROUGE-L is computed as
\begin{equation} 
	ROUGE \textendash L = \frac{\left( 1 + \gamma^2 \right) R_{LCS} P_{LCS} }{R_{LCS} + \gamma^2 P_{LCS}}
\end{equation}

$R_{LCS}$ is the ratio of LCS length to reference answer length, namely recall
\begin{equation} 
	R_{LCS} = \frac{LCS\left( c, r \right)}{|r|}
\end{equation}
where $c$ and $r$ represent the candidate and reference answer. 

$P_{LCS}$ is the ratio of LCS length to candidate answer length, namely precision.
\begin{equation} 
	P_{LCS} = \frac{LCS\left( c, r \right)}{|c|}
\end{equation}

For multiple gold answers, the maximum $R_{LCS}$ and $P_{LCS}$ are selected to compute ROUGE-L. Overall ROUGE-L on the dataset is defined as the average ROUGE value of each sample. 

Like our adaptations on BLEU, we integrate additional bonus items into $R_{LCS}$ and $P_{LCS}$. For yes-no answers, if $r$ and $c$ have the same opinion label, we add $\alpha LCS(r, c)$ to the numerator and denominator of $R_{LCS}$ and $P_{LCS}$. If participant judges opinions and the judgement is correct, the precision and recall will both increase, as is shown in example~\ref{exam3}.

{\small \begin{example}[Adapted ROUGE-L for yes-no answer]
	\label{exam3}
	\textbf{Question}: Is skipping rope an aerobic exercise?
	
	\noindent \textbf{Predicted answer}: [Yes] Skipping rope is an aerobic exercise. 
	
	\noindent \textbf{Gold answer 1}: [Yes] Skipping rope is a kind of aerobic exercise with low intensity.
	
	\noindent \textbf{Gold answer 2}: [Depends] Skipping rope can be regarded as an aerobic exercise only when skipping for a long time.
	
	\noindent \textbf{LCS length}: 6
	
	\noindent \textbf{LCS length for bonus}: 6 (LCS to gold answer 1)
	
	\noindent \textbf{Adapted $P_{LCS}$ ($\alpha=1.0$)}: (6 + 6) / (7 + 6) = 0.92

	\noindent \textbf{Adapted $R_{LCS}$ ($\alpha=1.0$)}: (6 + 6) / (12 + 6) = 0.67
	
	\noindent \textbf{Vanilla ROUGE-L\footnote{For simplicity, in this section we compute harmonic average to get ROUGE-L.}}: 0.59
	
	\noindent \textbf{Adapted ROUGE-L}: 0.78

\end{example}}

For entity answers, the bonus attached to the numerator and denominator of $R_{LCS}$ and $P_{LCS}$ is given as $\beta \sum_{e \in entities}{length(e) * I(e \subseteq c)}$, indicating the length sum of gold entities appearing in candidate answer. An example is given below.

{\small \begin{example}[Adapted ROUGE-L for entity answer]
	\label{exam4}
	\textbf{Question}: How long did it take for Qin Dynasty to unify China?

	\noindent \textbf{Predicted answer}: Qin unified China in 221 BC after the war against other kingdoms which lasted ten years.
	
	\noindent \textbf{Gold answer}: Qin unified China in ten years, from 230 BC to 221 BC.
	
	\noindent \textbf{Gold entity set}: ten years, 230 BC, 221 BC
	
	\noindent \textbf{LCS length}: 7
	
	\noindent \textbf{Entity length sum for bonus}: 4
	
	\noindent \textbf{Adapted $P_{LCS}$ ($\alpha=1.0$)}: (7 + 4) / (17 + 4) = 0.52
	
	\noindent \textbf{Adapted $R_{LCS}$ ($\alpha=1.0$)}: (7 + 4) / (14 + 4) = 0.61
	
	\noindent \textbf{Vanilla ROUGE-L}: 0.45	
	
	\noindent \textbf{Adapted ROUGE-L}: 0.56
	
\end{example}}

With the help of bonus items, our adapted metrics give more preference to correct yes-no and entity answers. For the yes-no question in example~\ref{exam1}~\&~\ref{exam3}, a trivial extracted answer may occur as \textit{``exercise with low intensity''} which does not contain yes-no opinion. The adapted ROUGE-L can better distinguish it from the correct one we give. With $\alpha$ set to 1.0, the right answer can achieve 0.28 higher point over the trivial one. When using vanilla ROUGE-L, the advantage narrows to only 0.09. For the entity question in example~\ref{exam2}~\&~\ref{exam4}, we consider a shorter candidate answer \textit{``Qin unified China in 221 BC after the war against other kingdoms''}. This answer lacks key information and should be assigned a lower score. However, this answer is preferred under vanilla ROUGE-L compared with the longer candidate in example~\ref{exam4} (0.53 vs 0.45). This problem will be rectified if the adapted ROUGE-L is employed with $\beta > 2.6$.

\section{Statistical Analysis}

To demonstrate the effectiveness of our adaptations, we measure the correlation of our metrics with human judgment quantitatively in comparison with original ROUGE-L and BLEU. 500 questions are sampled from DuReader, which cover yes-no, entity and description question types. We collect predicted answers to these questions from the submissions of 5 different MRC systems in MRC2018 Challenge\footnote{\href{http://mrc2018.cipsc.org.cn/}{http://mrc2018.cipsc.org.cn/}}. Generated opinion labels are attached with yes-no candidate answers, which is the common case in DuReader evaluation. The human judgment of these candidate answers is obtained by assigning 2 annotators to give the 1-5 score on each candidate. The overall human judgment score is defined as the average of scores across the questions. The correlation is analysed on both single question type and all the types. Meanwhile, the performances of these metrics are compared on both single question and overall score levels. The details of statistical analysis are given below.

\subsection{Human Judgment}

The samples we select include 201 yes-no, 201 entity and 98 description questions, with a total of 2500 candidate answers. The criterion of manual scoring is mainly based on whether the answer satisfies the demand of question, the coverage of key-points and answer conciseness. In detail, annotators give 1-5 scores according to the following guideline:

\begin{itemize*}
	\item \textbf{5-score}: perfectly answer the question with little redundant information
	\item \textbf{4-score}: sufficiently answer the question with unvital missing or some redundancy
	\item \textbf{3-score}: the answer is a little insufficient, such as only giving opinion without supporting context in yes-no answer
	\item \textbf{2-score}: vital missing or error exists
	\item \textbf{1-score}: totally irrelevant
\end{itemize*}

We follow the notion of \newcite{dang2007overview}, which emphasizes the coverage of vital key-point in the answer. Annotators are asked to treat yes-or-no opinion and important supporting information for yes-no questions and gold entities for entity questions as nuggets and put more weight on them for scoring.

To ensure the quality and credibility of the human judgment, we measure the argeement between the 2 annotators. Table~\ref{cor-human} shows the Pearson correlation coefficients for each question type and on overall. 

\begin{table}[!htbp]
	\small
	\begin{center}
		\begin{tabular}{c|c|c|c|c}
			\hline  & Yes-No & Entity & Description & Overall\\ \hline
			\textbf{PCC\footnotemark[4]} & 0.878 & 0.906 & 0.870 & 0.891\\
			\hline
		\end{tabular}
	\end{center}
	\caption{\label{cor-human}Pearson correlation coefficients (\textbf{PCC}) between annotators.}
\end{table}

We can see the annotators achieve high agreement on candidate judgment, which indicates the practicability of our scoring criterion and the reliability of the human annotation.\footnotetext[4]{All the PCCs are significant in t-test with p-value $<$ 0.05.}

\subsection{Effectiveness of Adaptations}

The correlation between automatic and manual evaluation metrics is calculated on both single question and overall score levels. On single question level, each candidate answer is taken as a sample to be scored by the two metrics and score pairs are collected across all the samples to compute PCC. On the overall level, predicted answers to 30 sampled questions for an MRC system are scored together and the resulting automatic and human overall score pairs are utilized for the calculation of PCCs. The sampling is performed 100 times and 5 systems are sampled the same questions each time. Hence each overall level PCC is computed on 500 samples.

In practice, we use cumulative BLEU-4 as the implementation of BLEU, which follows the official benchmark of DuReader. For ROUGE-L, $\gamma$ is set to 1.2 since we think the precision and recall are both of importance. The mean score given by 2 annotators are used to represent human judgment. For our adapted metrics, we set the weight of yes-no bonus $\alpha$ to 2.0 and that of entity bonus $\beta$ to 1.0.

On single question level, the pearson correlation coefficients between automatic metrics and human judgment are given in Table~\ref{cor-comp}. The adapted ROUGE-L achieves best performance on correlation to human judgment, both on single yes-no or entity question type and on overall. 

\begin{table}[!htbp]
	\small
	\begin{center}
		\begin{tabular}{c|c|c|c}
			\hline & Yes-No & Entity & Overall\\ \hline
			\textbf{Adapted ROUGE-L} & \textbf{0.540} & \textbf{0.620} & \textbf{0.570}\\
			\textbf{ROUGE-L} & 0.493 & 0.491 & 0.504\\
			\textbf{Adapted BLEU-4} & 0.478 & 0.469 & 0.481\\
			\textbf{BLEU-4} & 0.459 & 0.397 & 0.450\\
			\hline
		\end{tabular}
	\end{center}
	\caption{\label{cor-comp}PCCs between various automatic metrics and human judgment for different question types on single question level.}
\end{table}

Our adaptations bring substantial gain on PCCs for both ROUGE-L and BLEU-4 on single question level. To check the significance of these results, we follow the paired bootstrap resampling test mentioned in \newcite{koehn2004statistical}. For a pair of metrics, samples are bootstrapped 100 times and in each time the PCCs are recomputed and compared. For both ROUGE-L and BLEU-4, the paired test between original and adapted versions are performed on yes-no, entity and overall sets. In all the 6 tests, the adapted metric shows significant better performance than the original one.

We also calculate PCCs between automatic and human metrics on overall score level. The results are shown in Table~\ref{cor-comp-overall}. Similar to single question level, adapted ROUGE-L still gains the highest correlation to human overall judgment. In this task, we notice that ROUGE is much more effective than BLEU, which may reflect the importance of recall in MRC evaluation. For the comparison between adapted and vanilla metrics, adapted ROUGE-L performs better than vanilla version on every question type. However, our adapted BLEU-4 only works better on evaluating entity answers, which is different from the result on single question level. We think it may be due to the peculiar way BLEU employs to get overall score for multiple questions, which was discussed as the ``decomposability'' problem of BLEU in \newcite{chiang2008decomposability}. This issue will be explored in our future work.

\begin{table}[!htbp]
	\small
	\begin{center}
		\begin{tabular}{c|c|c|c}
			\hline & Yes-No & Entity & Overall\\ \hline
			\textbf{Adapted ROUGE-L} & \textbf{0.702} & \textbf{0.884} & \textbf{0.792}\\
			\textbf{ROUGE-L} & 0.664 & 0.839 & 0.760\\
			\textbf{Adapted BLEU-4} & 0.536 & 0.686 & 0.646\\
			\textbf{BLEU-4} & 0.571 & 0.668 & 0.681\\
			\hline
		\end{tabular}
	\end{center}
	\caption{\label{cor-comp-overall}PCCs between various automatic metrics and human judgment for different question types on overall score level.}
\end{table}

\subsection{Impacts of Bonus Weights}

\begin{figure}[t]
	\begin{center}
		\includegraphics[height=15em, width=20em]{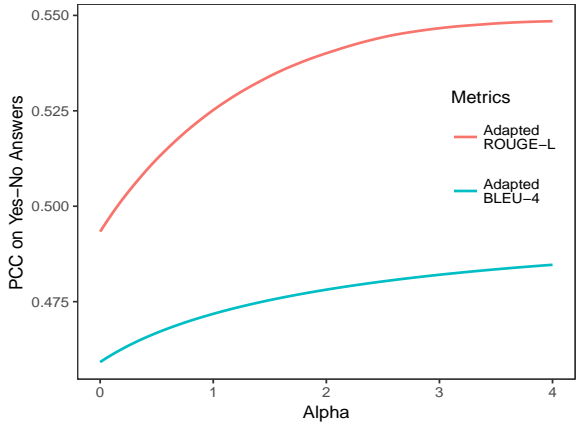}
	\end{center}
	\caption{\label{changinga}The PCC on yes-no answers w.r.t the variation of $\alpha$.}
\end{figure}

\begin{figure}[t]
	\begin{center}
		\includegraphics[height=15em, width=20em]{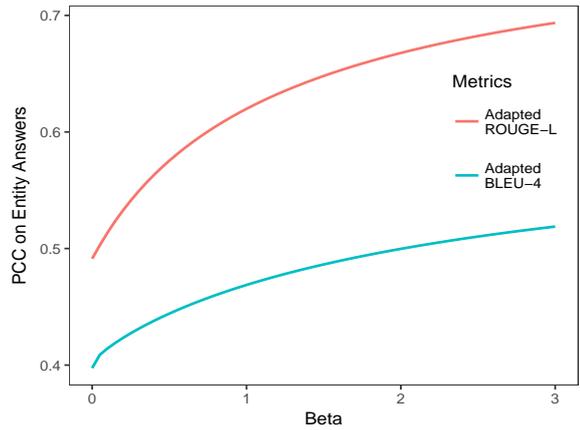}
	\end{center}
	\caption{\label{changingb}The PCC on entity answers w.r.t the variation of $\beta$.}
\end{figure}

We further inspect the impact of bonus weights on metric performance. In Figure~\ref{changinga}, the value of yes-no bonus weight $\alpha$ is changed with entity bonus weight $\beta$ fixed to 1.0. The single-question level PCCs of adapted BLEU-4 and ROUGE-L on yes-no answers are plotted w.r.t the variation of $\alpha$. We can see that the introduction of yes-no bonus brings positive effect for BLEU and ROUGE. Meanwhile, the PCCs of these metrics increase with $\alpha$ monotonically. 

Similarly, Figure~\ref{changingb} shows the single-question level PCCs of adapted BLEU-4 and ROUGE-L on entity answers w.r.t the variation of $\beta$, in which $\alpha$ is set to $2.0$. The effect of entity bonus is also positive and increases with $\beta$ monotonically. In future work, we will further look into the issue of selecting proper bonus weights.

\section{Conclusion}

For question answering MRC tasks, automatic evaluation metrics are commonly based on measuring lexical overlap, such as BLEU and ROUGE. However, in some cases, we notice that these automatic evaluation metrics may be biased from human judgment, especially for yes-no and entity questions. We think it may mislead the development of real scene MRC systems.

In this paper, we propose some adaptations to ROUGE and BLEU metrics for better evaluating yes-no and entity answers. Two bonus terms are introduced into the computation of original metrics. These terms are also based on lexical overlap. The statistical analysis shows that our adaptations achieve higher correlation to human judgment compared with original ROUGE-L and BLEU, proving the effectiveness of our methodology. In the future, our work will cover more question types and more MRC datasets. We hope our exploration can bring more research attention to the design of MRC evaluation metrics.

\section{Acknowledgement}
We thank the anonymous reviewers for their insightful comments on this paper. 
This work was partially supported by National Natural Science
Foundation of China (61572049) and Baidu-Peking University Joint Project.

\bibliography{acl2018}
\bibliographystyle{acl_natbib}

\end{document}